%% file: main.tex
\documentclass[letterpaper]{article}
\usepackage[preprint]{aaai2027}
\usepackage[hyphens]{url}
\usepackage{graphicx}
\usepackage[numbers,sort&compress]{natbib}
\usepackage{xcolor}
\definecolor{DIVARefBlue}{RGB}{20,55,105}
\AtBeginDocument{%
  \let\DIVAOriginalCite\cite
  \renewcommand{\cite}[1]{\textcolor{DIVARefBlue}{\DIVAOriginalCite{#1}}}%
}
\usepackage{caption}
\usepackage{booktabs}
\usepackage{amsmath}
\usepackage{amssymb}
\usepackage{multirow}
\title{DIVA: Dual-Space Intent-Aware Visual Attenuation for Vision-Language-Action Policies}
\author{Kaixi Feng, Guoheng Sun, Ziyao Wang, Yexiao He, Zheyu Shen, Ang Li}
\affiliations{University of Maryland, College Park}

\begin{document}
\maketitle

\begin{abstract}
Vision-language-action (VLA) policies feed dense visual patch tokens to a language-action backbone, preserving scene context without explicitly regulating each token's influence on policy computation. We introduce DIVA, a Dual-Space Intent-Aware Visual Attenuation module with an anchor-then-attenuate design. DIVA combines task intent with local visual evidence to estimate patch-wise relevance anchors and applies them in two complementary spaces: reweighting projected visual tokens before backbone entry and persistently attenuating low-relevance visual states within the backbone. DIVA preserves the full visual token sequence and requires no external grounding supervision. On LIBERO, DIVA improves OpenVLA-OFT from 96.6\% to 98.0\% average success and raises its zero-shot LIBERO-Plus score from 69.6 to 72.6. Real-world experiments further show consistent gains under task-irrelevant visual perturbations, supporting the robustness of intent-aware visual attenuation beyond simulation.
\end{abstract}

\section{Introduction}

Vision-language-action (VLA) policies have become a promising foundation for language-conditioned robot manipulation \cite{openx2024rtx,li2023roboflamingo}. By adapting large vision-language backbones for robot action prediction, recent models can map visual observations and natural-language instructions to actions across diverse tasks \cite{zitkovich2023rt2,kim2025openvla,black2024pi0}. A common design is to encode one or more camera views into dense visual patch tokens, project them into a language-action backbone, and let the backbone transform visual-language representations into robot actions \cite{liu2024rdt,qu2025spatialvla}.

This dense-token interface preserves rich scene information, but offers no explicit mechanism for regulating how strongly each visual token contributes to policy computation. Consequently, task-irrelevant objects may exert undue influence on the policy and impair action prediction. This raises a central question: \textit{how can a VLA policy attenuate weakly relevant visual evidence while preserving the scene context needed for reliable manipulation?}

Soft attenuation offers a non-destructive way to meet this objective. Rather than removing weakly relevant tokens, it reduces their contribution while keeping the complete visual sequence available to the policy. This distinction matters because evidence peripheral to the current instruction may still support spatial reasoning, obstacle awareness, or safe interaction. The goal is therefore not a binary partition of the scene, but a graded allocation of influence across visual tokens.

Soft attenuation provides the general strategy, but answering this question in practice requires resolving two challenges. The first is \emph{how to estimate visual-token relevance}. High-level task intent indicates what the policy should seek, but remains spatially ambiguous. Low-level visual evidence identifies local structures such as handles, contact regions, object boundaries, and free space, but may emphasize salient yet task-irrelevant content. Human gaze is jointly guided by top-down task intent and bottom-up visual evidence \cite{wolfe1994guided,corbetta2002control}. Inspired by this process, we combine high-level task intent with low-level visual evidence to construct patch-wise relevance anchors.

The second is \emph{where attenuation should act once the anchors are established}. Input space provides a natural starting point: projected visual tokens retain explicit patch correspondence, allowing the anchors to set the initial strength of each token before backbone processing. However, this one-time adjustment may be diluted by subsequent attention, feed-forward transformations, and residual updates. Deeper in the backbone, visual hidden states lie closer to action computation, making them a natural place to sustain attenuation. These complementary roles motivate a dual-space design: input-space attenuation initializes patch-aligned representation strength, while latent-space attenuation maintains it throughout policy computation.

These two answers lead directly to DIVA, a Dual-Space Intent-Aware Visual Attenuation module for VLA policies. DIVA follows an anchor-then-attenuate design. It first combines high-level task intent with low-level visual evidence to estimate patch-wise relevance anchors. It then applies the same anchors in both input and latent spaces, reweighting projected visual tokens before backbone entry and attenuating low-relevance visual hidden states across layers. This design preserves the full visual token sequence and requires no external grounding labels.

We instantiate DIVA on OpenVLA-OFT and evaluate it on LIBERO, zero-shot LIBERO-Plus perturbations, and real-world manipulation tasks \cite{kim2025oft,liu2023libero,fei2025liberoplus}. DIVA improves the baseline across these settings, while controlled ablations validate the complementary roles of high- and low-level evidence in anchoring and of input- and latent-space attenuation.

Our contributions are summarized as follows:
\begin{itemize}
    \item We identify a key limitation of dense-token VLA policies: they lack an explicit mechanism to control the influence of visual tokens according to task relevance.
    
    \item We introduce DIVA, an anchor-then-attenuate framework that combines high-level task intent with low-level visual evidence to construct patch-wise relevance anchors and applies them in both input and latent spaces.
    
    \item We evaluate DIVA on LIBERO, zero-shot LIBERO-Plus, and real-world manipulation tasks, demonstrating consistent gains under standard and perturbed conditions. Controlled ablations verify the contributions of Grounded Text Pooling, intent-aware relevance anchoring, and dual-space attenuation.
\end{itemize}

\section{Related Work}

\begin{figure}[t]
    \centering
    \includegraphics[width=0.98\columnwidth]{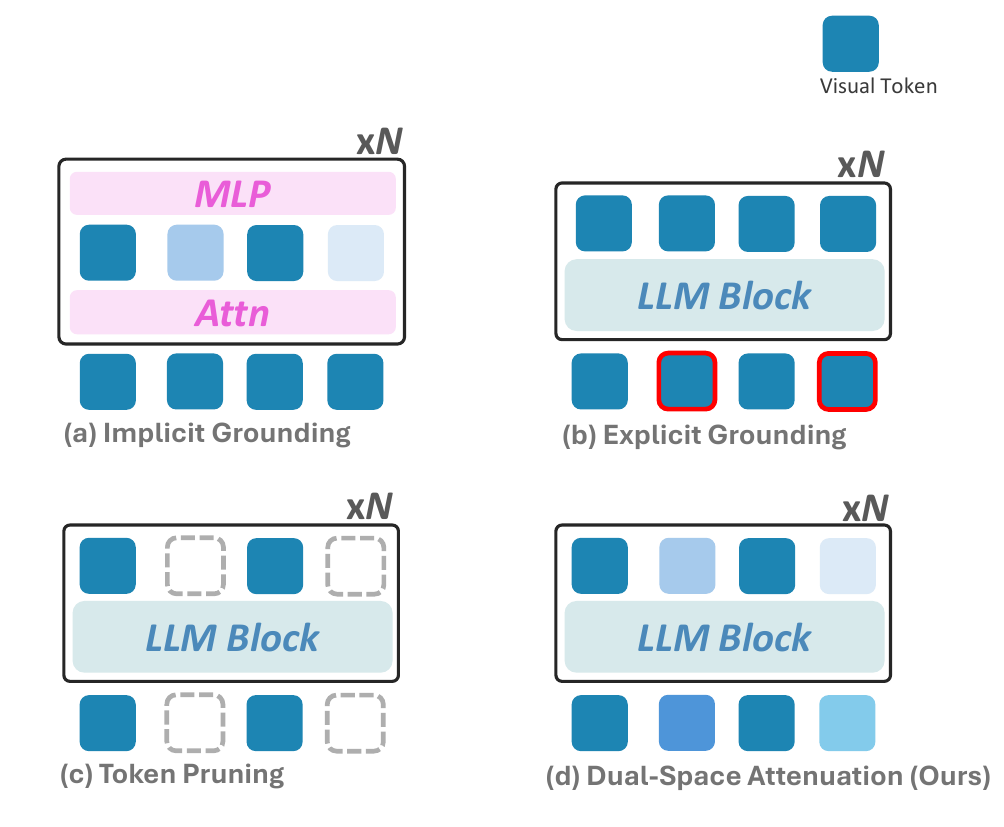}
    \caption{
    Conceptual comparison with related visual relevance mechanisms in VLA policies.
    Implicit grounding methods guide visual representations through internal objectives or attention-side signals;
    explicit grounding methods localize task-relevant visual regions;
    token pruning methods remove weakly relevant tokens.
    In contrast, DIVA preserves the full visual token sequence and softly attenuates visual representations in both input space and latent space.
    }
    \label{fig:related_work}
\end{figure}

\paragraph{Explicit grounding and input-side localization.}
Explicit visual grounding supplies spatial cues before action prediction, commonly through external detectors, segmenters, masks, boxes, cropped regions, or visual prompts \cite{zheng2024tracevla}. RoboGround uses grounding masks as intermediate observations \cite{huang2025roboground}, while other grounding-oriented VLAs generate object boxes, end-effector states, or grasp cues before predicting actions \cite{zawalski2024ecot,deng2025graspvla}. Such cues are interpretable and can reduce ambiguity in object-centric manipulation. However, they rely on external models, additional supervision, or generated spatial targets, and their effectiveness depends on localization quality. Incorrect masks or overly narrow crops may discard scene context needed for spatial reasoning and collision avoidance. DIVA avoids external grounding targets and derives relevance anchors from the policy's own language and visual representations.

\paragraph{Implicit grounding and attention-side visual modulation.}
Related approaches guide visual focus through attention mechanisms or auxiliary supervision. ReconVLA reconstructs gaze regions for manipulated objects \cite{song2025reconvla}, gaze-regularized VLAs align internal attention with gaze-derived distributions \cite{pani2026gazereg}, and AVA-VLA and FocusVLA modulate attention using recurrent states or task-conditioned visual signals \cite{xiao2025avavla,zhang2026focusvla}. Their control acts locally through attention and may be reshaped by subsequent feed-forward and residual updates, while gaze-supervised variants require additional annotations. In contrast, DIVA directly regulates representation strength by using patch-wise relevance anchors to reweight projected visual tokens before backbone entry and attenuate the corresponding visual hidden states across layers, thereby sustaining attenuation throughout policy computation.

\paragraph{Visual token pruning and selection.}
Visual token pruning reduces redundancy by retaining only selected tokens \cite{ryoo2021tokenlearner}. LightVLA uses differentiable pruning for efficient inference \cite{jiang2025lightvla}, DTP removes distracting image tokens \cite{li2026dtp}, and ADP and related sparsification methods introduce action-aware or text-guided selection inside VLA backbones \cite{pei2025adp,wang2026etavla}. Although efficient, hard keep-or-remove decisions may discard weakly salient regions that still encode spatial layout, object relations, obstacles, or target context. DIVA therefore takes a non-destructive alternative: it preserves the full visual sequence and softly attenuates low-relevance representations rather than deleting them.

\section{Method}
\label{sec:method}

\subsection{Overview}

DIVA is designed to control the strength of visual evidence in a token-based vision-language-action (VLA) policy. Given an instruction and visual observations, the base policy encodes images into projected visual tokens and feeds them, together with instruction tokens, into a language-action backbone. Let
$V=\{v_i\}_{i=1}^{N}$ denote the projected visual tokens and
$H=\{h_j\}_{j=1}^{T}$ denote the instruction-token embeddings. 
DIVA operates on the projected visual-token sequence and preserves its token layout throughout the policy.

\begin{figure*}[t]
\centering
\includegraphics[width=0.96\textwidth]{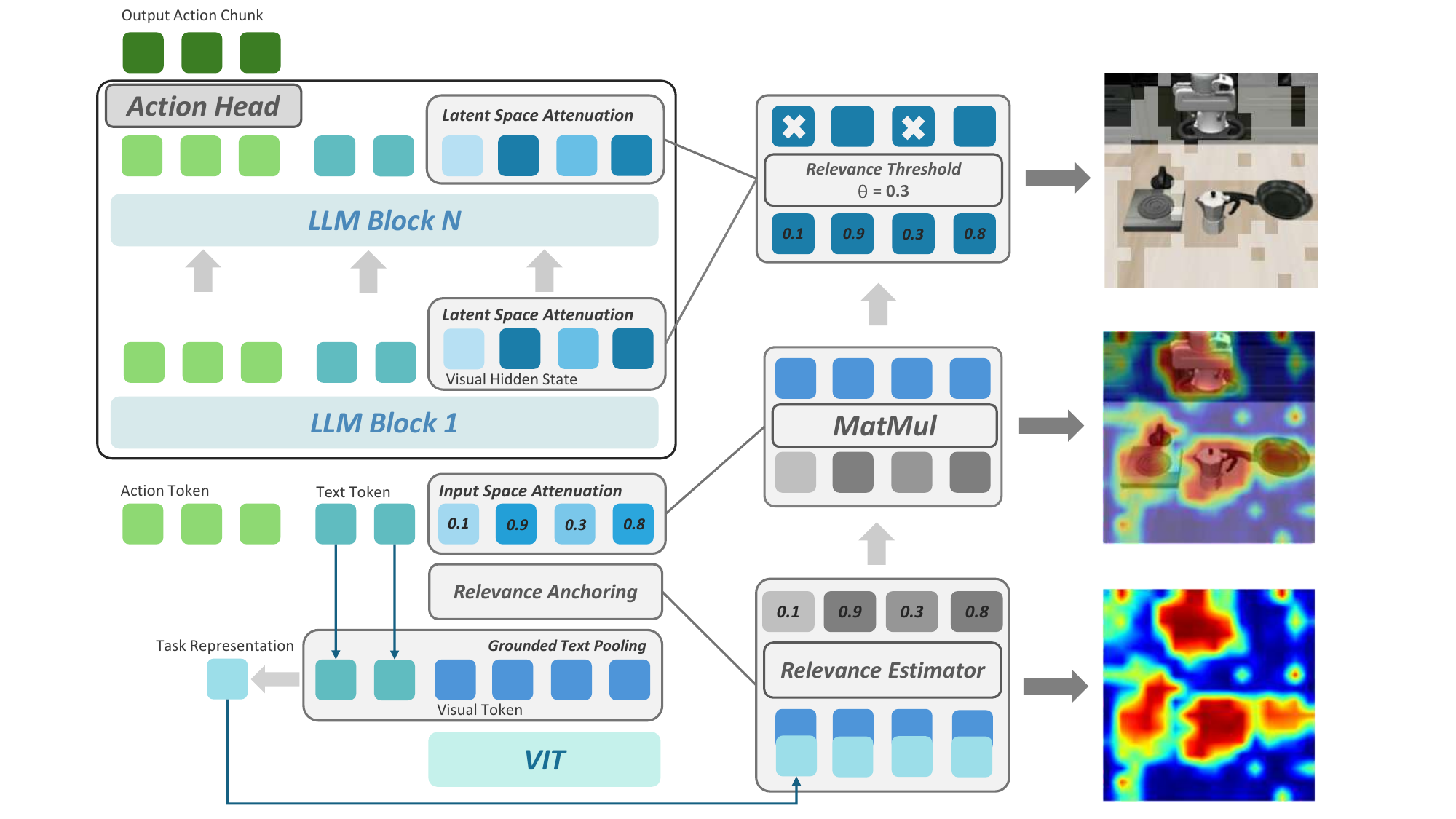}
\caption{
Overview of DIVA on the task \emph{``turn on the stove and put the moka pot on it.''}
DIVA first obtains a compact task representation through Grounded Text Pooling and estimates patch-wise relevance anchors from task-conditioned visual tokens. 
The same anchors are then used for dual-space visual attenuation: explicit input-space attenuation reweights projected visual tokens before backbone input, while implicit latent-space attenuation persistently weakens low-relevance visual hidden states across backbone layers.
}
\label{fig:method_overview}
\end{figure*}

Following its anchor-then-attenuate design, DIVA estimates each patch-wise relevance anchor $\alpha_i$ from task intent and local visual evidence, reweights projected visual tokens before backbone entry, and reuses the anchors to attenuate low-relevance hidden states across layers.

\subsection{Grounded Text Pooling}

Intent-aware relevance anchoring requires a task representation that tells the policy what to seek in the current observation. Instruction tokens are heterogeneous: object names, target descriptions, and relation words specify the executable intent, whereas generic function words carry little visual specificity. Uniform mean pooling ignores this distinction and, because it is independent of the observation, cannot determine which instruction concepts are visually instantiated in the current scene. We therefore introduce Grounded Text Pooling (GTP), which uses localized visual evidence to determine how instruction tokens contribute to a compact, observation-conditioned representation of task intent.

Given instruction-token embeddings $H=\{h_j\}_{j=1}^{T}$ and projected visual tokens $V=\{v_i\}_{i=1}^{N}$, we first measure the compatibility between every instruction token and visual patch:
\begin{equation}
    s_{j,i} =
    \frac{h_j^{\top}v_i}{\|h_j\|_2\|v_i\|_2}.
\end{equation}
For each instruction token $h_j$, the vector $s_j=\{s_{j,i}\}_{i=1}^{N}$ forms a response map over visual patches. A visually instantiated concept need not match the image globally; instead, it should receive distinctive support from a subset of patches depicting its referent, target, or relation. Generic function words carry little task-specific semantics. Treating all instruction tokens equally would therefore dilute the compact task summary needed for relevance anchoring. We therefore assess each instruction token by its support from localized visual patches rather than by global image-level compatibility \cite{yao2021filip,mukhoti2023pacl}.

To summarize localized visual support, we compute a soft local response:
\begin{equation}
    a_{j,i} =
    \frac{\exp(s_{j,i}/\tau_v)}
    {\sum_{k=1}^{N}\exp(s_{j,k}/\tau_v)},
    \qquad
    e_j = \sum_{i=1}^{N} a_{j,i}s_{j,i},
\end{equation}
where $\tau_v$ controls the sharpness of visual selection. The term $e_j$ emphasizes the patches most strongly aligned with instruction token $h_j$.

However, a high local response alone does not imply that a token provides discriminative visual evidence. A token that is broadly compatible with the entire image may also obtain a large response, even though it does not identify any particular part of the scene. We therefore contrast the soft local response with a uniform visual baseline:
\begin{equation}
    \bar{s}_j =
    \frac{1}{N}\sum_{i=1}^{N}s_{j,i},
    \qquad
    c_j = e_j - \bar{s}_j .
\end{equation}
The score $c_j$ measures the gain from localized visual selection over global image-level compatibility. Diffuse responses produce little gain, whereas tokens supported by a distinctive subset of patches receive larger scores. Thus, $c_j$ captures whether an instruction token is selectively grounded in the current observation, rather than merely compatible with it overall.

We then convert these grounded scores into pooling weights:
\begin{equation}
    \omega_j =
    \frac{\exp(c_j/\tau_t)}
    {\sum_{k=1}^{T}\exp(c_k/\tau_t)},
    \qquad
    t = \sum_{j=1}^{T}\omega_j h_j,
\end{equation}
where $\tau_t$ controls the sharpness of text-token pooling. The resulting vector $t$ emphasizes instruction semantics with localized, discriminative support in the current scene and serves as the high-level task-intent representation for relevance anchoring.

\subsection{Intent-Aware Visual Relevance Anchoring}
Intent-aware anchoring follows human goal-directed vision, where gaze combines top-down task intent with bottom-up visual evidence \cite{wolfe1994guided,corbetta2002control}. Task intent specifies what to seek---such as an object, target, or relation---while patch-level tokens indicate where compatible cues, such as boundaries, handles, and contact regions, appear. Intent is spatially ambiguous on its own, whereas local evidence may favor salient but task-irrelevant regions. DIVA therefore combines both signals to estimate patch-wise relevance.

Let $t$ denote the task-intent representation produced by Grounded Text Pooling. DIVA estimates relevance anchors in the projected visual-token space, where explicit patch correspondence remains available before backbone entry.

For each visual token, we concatenate its low-level visual representation with the high-level task intent:
\begin{equation}
    z_i = [v_i; t],
    \qquad
    Z=\{z_i\}_{i=1}^{N}.
\end{equation}
A lightweight relevance estimator then reasons over the task-conditioned visual sequence:
\begin{equation}
    \widetilde{Z}=\mathcal{R}_{\phi}(Z),
\end{equation}
where $\mathcal{R}_{\phi}$ is implemented as a lightweight self-attention block that contextualizes the task-conditioned patch sequence. Rather than scoring each patch in isolation, it allows a patch to assess its relevance relative to other scene elements, capturing dependencies among manipulated objects, targets, and surrounding geometry. Each output $\widetilde{z}_i$ remains aligned with the corresponding input patch and summarizes both its local appearance and inter-patch context.

Each conditioned visual token is mapped to a scalar logit:
\begin{equation}
    r_i=g_{\phi}(\widetilde{z}_i),
\end{equation}
and the visual relevance anchor is obtained with a sigmoid head:
\begin{equation}
    \alpha_i=\sigma(r_i),
    \qquad
    \alpha_i\in[0,1].
\end{equation}
Task-relevant objects and interaction regions often span multiple visual patches. Independent sigmoid predictions allow every patch belonging to such a region to receive a high relevance score, preserving the full spatial extent of task-relevant evidence.

\subsection{Dual-Space Visual Attenuation}

After relevance anchoring, DIVA applies the same anchors in two spaces. Explicit input-space attenuation reweights patch-aligned projected tokens once before backbone entry, controlling their initial strength. Implicit latent-space attenuation repeatedly modulates the corresponding visual hidden states as they evolve across layers. The shared anchors therefore connect explicit input control with persistent latent-state attenuation.

\paragraph{Explicit input-space attenuation.}
The explicit path directly reweights projected visual tokens:
\begin{equation}
    v_i' = \alpha_i v_i .
\end{equation}
Since $v_i$ still has a patch-indexed interpretation, this operation provides an explicit relevance-aware input sequence while preserving the original visual token layout. The same anchor $\alpha_i$ remains associated with the corresponding visual-token position and is reused for latent-space attenuation.

\paragraph{Implicit latent-space attenuation.}
Input-space attenuation controls the visual evidence entering the backbone, but it is applied only once. As attention, feed-forward transformations, and residual updates progressively reshape the representations, its effect may fade. Latent space provides a complementary intervention point because visual states continue to evolve across backbone layers and lie closer to action computation. DIVA therefore carries the same relevance anchors into the latent space and reapplies them to these evolving visual states across layers.

Let $m_i^{\ell}$ denote the hidden state corresponding to the $i$-th visual token at layer $\ell$. Given its relevance anchor $\alpha_i$, DIVA updates the state as
\begin{equation}
    \widetilde{m}_i^{\ell} =
    \begin{cases}
    m_i^{\ell}, & \alpha_i \geq \theta, \\
    (1-\lambda)m_i^{\ell} + \lambda(\alpha_i m_i^{\ell}), & \alpha_i < \theta,
    \end{cases}
\end{equation}
where $\theta$ determines which states are attenuated and $\lambda\in[0,1]$ controls the attenuation strength. States above the threshold remain intact, protecting sufficiently relevant evidence from repeated weakening. Below the threshold, interpolation scales the state by $1-\lambda(1-\alpha_i)$, retaining a lower bound of $1-\lambda$ and preserving contextual information. We use $\theta=0.3$ and $\lambda=0.5$ throughout. The threshold only routes the latent attenuation path; the anchors remain trainable through input-space attenuation and the relevance budget loss.

Reapplying this rule across layers maintains attenuation as visual representations evolve. High-relevance states preserve their strength, while low-relevance states continue to provide reduced but nonzero contextual support. This produces persistent, non-destructive visual attenuation throughout the backbone.

\subsection{Training Objective}

DIVA is trained together with the original action imitation objective of the base VLA policy. Let $\mathcal{L}_{\mathrm{act}}$ denote this action prediction loss. In our OpenVLA-OFT instantiation \cite{kim2025oft}, the policy predicts a continuous action chunk $\hat{A}=\{\hat{a}_h\}_{h=1}^{H}$ for a ground-truth chunk $A=\{a_h\}_{h=1}^{H}$, and we use the standard L1 regression objective:
\begin{equation}
    \mathcal{L}_{\mathrm{act}}
    =
    \frac{1}{H}
    \sum_{h=1}^{H}
    \left\|
    \hat{a}_h-a_h
    \right\|_1 .
\end{equation}

Because relevance anchors are predicted independently, the anchoring module could assign high relevance to every visual token. We prevent this with a relevance-budget loss:
\begin{equation}
    \mathcal{L}_{\mathrm{budget}} =
    \left(
    \frac{1}{N}\sum_{i=1}^{N}\alpha_i - \rho
    \right)^2,
\end{equation}
where $\rho$ is the target average relevance. The final objective is
\begin{equation}
    \mathcal{L}=
    \mathcal{L}_{\mathrm{act}}
    +\eta\mathcal{L}_{\mathrm{budget}},
\end{equation}
where $\eta$ controls the strength of the budget regularization. We set $\rho=0.4$ and $\eta=0.01$ in all experiments. The moderate budget $\rho=0.4$ avoids both near-uniform relevance and overly sparse visual evidence. Together with $\theta=0.3$, it provides stable optimization while attenuating a substantial portion of low-relevance, task-irrelevant visual evidence. The budget encourages a compact visual evidence pathway while keeping DIVA grounding-supervision-free.

\section{Experiments}
\label{sec:experiments}

We evaluate DIVA on the standard LIBERO benchmark and the robustness-oriented LIBERO-Plus benchmark \cite{liu2023libero,fei2025liberoplus}. LIBERO covers four manipulation suites: Spatial, Object, Goal, and Long, while LIBERO-Plus evaluates seven controlled perturbations involving camera viewpoint, robot state, language, lighting, background texture, sensor noise, and object layout. Following the zero-shot protocol, DIVA is trained only on standard LIBERO demonstrations and evaluated on LIBERO-Plus without training or adaptation on perturbation data. We instantiate DIVA on OpenVLA-OFT \cite{kim2025oft} and compare it with representative policy and VLA baselines \cite{chi2024diffusionpolicy,zheng2024tracevla,team2024octo,kim2025openvla,hou2025dita,zhao2025cotvla,pertsch2025fast,black2024pi0,wang2025univla,jiang2025lightvla,pei2025adp}. Simulation experiments use 8 NVIDIA L20 GPUs; additional implementation details are provided in Appendix~\ref{sec:implementation}.

\paragraph{Evaluation protocol.}
We train one policy jointly on all 40 LIBERO tasks. Each task is evaluated for 50 episodes from benchmark-provided initial states, yielding 500 trials per suite and 2,000 trials in total. Success is determined by the sparse task-completion predicate. Training uses a random crop covering 90\% of the image area, and evaluation uses the corresponding center crop. Both camera views are resized to $224\times224$. On LIBERO-Plus, each of the 10,030 variants is executed once per evaluation run. The total score is weighted by the number of variants, rather than averaging the seven factor scores.

\paragraph{Training protocol.}
The baseline, DIVA, and all ablation variants use matched total optimization budgets: 160K steps in simulation and 30K steps on the physical robot. DIVA initializes from a task-trained OpenVLA-OFT checkpoint and activates its modules only during the final 10K steps. Simulation training uses LoRA rank 32, an action chunk length of 8, and an effective batch size of 64. Physical-robot training uses an action chunk length of 25 and a global batch size of 32 on four NVIDIA L20 GPUs. Stage-specific optimizer settings are given in the appendix.

\begin{table}[b]
\centering
\small
\setlength{\tabcolsep}{2.2pt}
\begin{tabular}{@{}lccccc@{}}
\toprule
\textbf{Method} & \textbf{Spatial} & \textbf{Object} & \textbf{Goal} & \textbf{Long} & \textbf{Avg.} \\
\midrule
Diffusion Policy & 78.3 & 92.5 & 68.3 & 50.5 & 72.4 \\
TraceVLA & 84.6 & 85.2 & 75.1 & 54.1 & 74.8 \\
Octo & 78.9 & 85.7 & 84.6 & 51.1 & 75.1 \\
OpenVLA & 84.7 & 88.4 & 79.2 & 53.7 & 76.5 \\
DiTA & 84.2 & 96.3 & 85.4 & 63.8 & 82.4 \\
CoT-VLA & 87.5 & 91.6 & 87.6 & 69.0 & 83.9 \\
$\pi_0$-FAST & 96.4 & 96.8 & 88.6 & 60.2 & 85.5 \\
$\pi_0$ & 96.8 & 98.8 & 95.8 & 85.2 & 94.2 \\
UniVLA & 96.5 & 96.8 & 95.6 & 92.0 & 95.2 \\
LightVLA & 98.4 & 98.4 & 98.2 & 94.6 & 97.4 \\
VLA-ADP & \textbf{99.4} & 98.0 & 96.4 & 91.2 & 96.3 \\
OpenVLA-OFT$^\dagger$ & 97.6 & 97.6 & 96.0 & 95.0 & 96.6 \\
\midrule
\textbf{DIVA (Ours)} & 99.0 & \textbf{99.4} & \textbf{98.4} & \textbf{95.2} & \textbf{98.0} \\
\bottomrule
\end{tabular}
\caption{
LIBERO success rates (\%) across four task suites.
OpenVLA-OFT$^\dagger$ is our evaluation of the official checkpoint.
}
\label{tab:libero_main}
\end{table}

\subsection{Evaluation Results}

\paragraph{Standard manipulation performance.}
Table~\ref{tab:libero_main} reports the LIBERO results. DIVA achieves a 98.0\% average success rate and outperforms OpenVLA-OFT across all four suites. The gains on Spatial and Object, which require identifying task-relevant objects and spatial configurations, support the effectiveness of intent-aware relevance modeling.

\begin{table*}[t]
\centering
\setlength{\tabcolsep}{3.0pt}
\begin{tabular*}{\textwidth}{@{\extracolsep{\fill}}lcccccccc@{}}
\toprule
\textbf{Method} & \textbf{Cam.} & \textbf{Robot} & \textbf{Lang.} & \textbf{Light} & \textbf{Backg.} & \textbf{Noise} & \textbf{Layout} & \textbf{Total} \\
\midrule
OpenVLA~\cite{kim2025openvla} & 0.8 & 3.5 & 23.0 & 8.1 & 34.8 & 15.2 & 28.5 & 15.6 \\
NORA~\cite{hung2025nora} & 2.2 & 37.0 & 65.1 & 45.7 & 58.6 & 12.8 & 62.1 & 39.0 \\
UniVLA~\cite{wang2025univla} & 1.8 & \textbf{46.2} & 69.6 & 69.0 & 81.0 & 21.2 & 31.9 & 42.9 \\
RIPT-VLA~\cite{tan2025riptvla} & 55.2 & 31.2 & 77.6 & 88.4 & 91.6 & 73.5 & 74.2 & 68.4 \\
OpenVLA-OFT~\cite{kim2025oft} & 56.4 & 31.9 & 79.5 & 88.7 & 93.3 & 75.8 & 74.2 & 69.6 \\
\midrule
\textbf{DIVA (Ours)} & \textbf{62.4} & 33.7 & \textbf{80.6} & \textbf{92.6} & \textbf{96.1} & \textbf{79.1} & \textbf{76.3} & \textbf{72.6} \\
\bottomrule
\end{tabular*}
\caption{
Zero-shot LIBERO-Plus success scores (\%) across perturbation factors. Total is task-weighted over all evaluation instances.
}
\label{tab:libero_plus}
\end{table*}

\paragraph{Robustness under perturbations.}
DIVA improves the task-weighted score of the official OpenVLA-OFT checkpoint from 69.6 to 72.6. The gains are consistent across all perturbation factors, with clear improvements on Camera, Light, Noise, Background, and Layout. These factors directly perturb the visual stream, supporting the role of dual-space attenuation in reducing the influence of unstable or low-relevance visual evidence. DIVA also improves under language perturbations, indicating that its relevance estimation remains effective under instruction variation. Over five seeds, DIVA obtains $98.00\pm0.39\%$ and $72.60\pm0.43\%$ on LIBERO and LIBERO-Plus, respectively (mean $\pm$ sample standard deviation).

\subsection{Real-World Experiments}

\paragraph{Experimental setup.}
We evaluate DIVA on a physical UFACTORY UF850 robot using synchronized third-person and wrist-mounted Intel RealSense observations. DIVA and OpenVLA-OFT are fine-tuned on the same VR-teleoperated demonstrations. Following a capability-oriented protocol, we organize the benchmark into three task suites: (1) \emph{Simple Pick-and-Place}, where the robot picks an object and places it into a bowl specified by color; (2) \emph{Long-Horizon Manipulation}, where the robot presses multiple colored buttons in the instructed order; and (3) \emph{Visually Grounded Manipulation}, where the policy must identify the correct cube or pair of cups among visually similar candidates before placing the selected cube into a cup or stacking the instructed cups.

For every task, we evaluate the standard workspace and two OOD conditions. \emph{Unseen Objects} adds distractors around task-relevant objects to test local interference, while \emph{Unseen Background} changes the tabletop to test global visual shifts. We run 25 trials per task--condition pair and report successful executions.

\begin{figure*}[t]
    \centering
    \includegraphics[width=\textwidth]{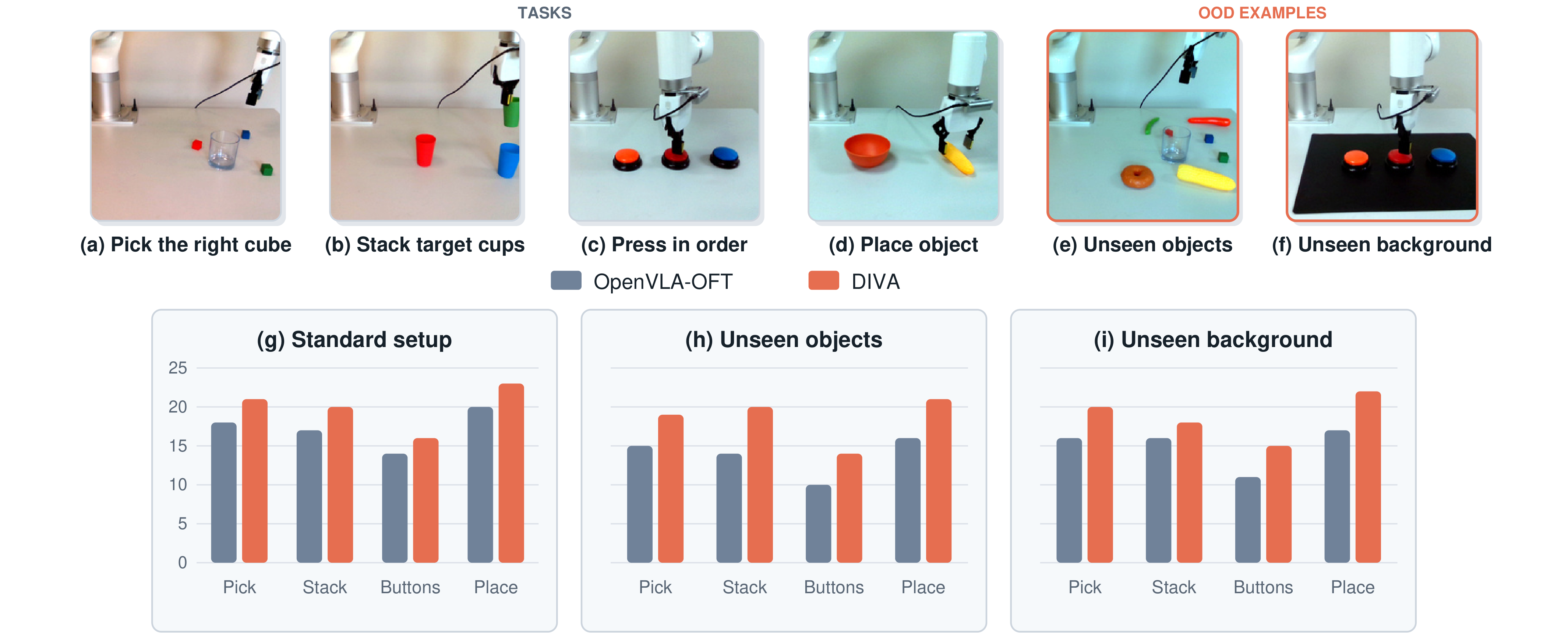}
    \caption{Real-world evaluation on the UF850 platform. Successful executions out of 25 physical trials for OpenVLA-OFT and DIVA under the standard setup, unseen objects, and an unseen background.}
    \label{fig:real_world}
\end{figure*}

\paragraph{Real-world results.}
Figure~\ref{fig:real_world} shows that DIVA completes more trials than OpenVLA-OFT in every evaluated task--condition pair. Stacking benefits most from unseen-object distractors, which require distinguishing the instructed cups from distractors. The pick-the-right-cube task shows a similar gain when grounding a referring expression to the correct object among visually similar alternatives.

DIVA also achieves higher success under both irrelevant-object distractors and background shifts. The sequential-button task is the most challenging, plausibly because the policy must infer task progress from visually similar intermediate states, whereas simple pick-and-place remains reliable across settings. Together, these results indicate that dual-space intent-aware attenuation improves both semantic object selection and robustness during physical deployment.

\subsection{Ablation Studies}

Table~\ref{tab:ablation} isolates task-conditioned anchoring and dual-space attenuation. The baseline and all ablation variants are trained independently under the same training budget; only the ablated component differs.

\begin{center}
\begin{minipage}{\columnwidth}
\centering
\small
\setlength{\tabcolsep}{1.4pt}
\begin{tabular}{@{}lccccc@{}}
\toprule
\multicolumn{6}{c}{\textit{(a) Relevance Anchoring}} \\
\midrule
\textbf{Strategy} & \textbf{Spatial} & \textbf{Object} & \textbf{Goal} & \textbf{Long} & \textbf{Avg.} \\
\midrule
Baseline & 97.6 & 97.6 & 96.0 & 94.8 & 96.5 \\
Visual-only & 91.8 & 98.2 & 97.2 & 94.2 & 95.4 \\
+ Mean Text Pooling & 99.0 & 99.2 & 97.0 & 94.6 & 97.5 \\
+ Grounded Text Pooling & \textbf{99.0} & \textbf{99.4} & \textbf{98.4} & \textbf{95.2} & \textbf{98.0} \\
\midrule
\multicolumn{6}{c}{\textit{(b) Dual-Space Attenuation}} \\
\midrule
\textbf{Design} & \textbf{Spatial} & \textbf{Object} & \textbf{Goal} & \textbf{Long} & \textbf{Avg.} \\
\midrule
Baseline & 97.6 & 97.6 & 96.0 & 94.8 & 96.5 \\
Explicit only & 97.8 & 99.0 & 97.4 & \textbf{95.4} & 97.4 \\
Implicit only & 98.4 & \textbf{99.6} & 97.2 & 94.6 & 97.5 \\
Dual-space attenuation & \textbf{99.0} & 99.4 & \textbf{98.4} & 95.2 & \textbf{98.0} \\
\bottomrule
\end{tabular}
\captionof{table}{
LIBERO ablations of (a) relevance anchoring and (b) dual-space attenuation.
}
\label{tab:ablation}
\end{minipage}
\end{center}

\paragraph{Relevance anchoring.}
Panel (a) directly tests the two signals used for intent-aware anchoring. Visual-only anchoring removes top-down task intent and substantially underperforms the task-conditioned variants, showing that local saliency alone cannot determine which visual evidence matters for a given instruction. Mean Text Pooling supplies a task-intent representation but aggregates instruction tokens uniformly. Grounded Text Pooling performs best by combining task intent with localized visual evidence, supporting the high--low complementarity behind DIVA's anchoring design.

\paragraph{Dual-space attenuation.}
Panel (b) tests whether both spaces are necessary. Explicit Only applies anchors before backbone input; Implicit Only applies them in its latent path. Both improve over the baseline but fall short of the full dual-space design. This supports the anchor-then-attenuate mechanism: explicit attenuation initializes a relevance-aware visual sequence, while implicit attenuation preserves this control through attention, feed-forward updates, and residual mixing.

\begin{figure*}[!t]
\centering
\small
\setlength{\tabcolsep}{1.0pt}
\begin{tabular}{c c c c c}
& \textbf{Step 0} & \textbf{Step 10} & \textbf{Step 20} & \textbf{Step 30} \\[-1pt]
\rotatebox{90}{\textbf{Third-person}}
&
\includegraphics[width=0.083\textwidth]{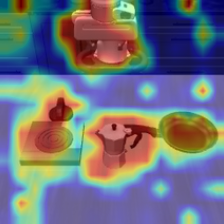}
\includegraphics[width=0.083\textwidth]{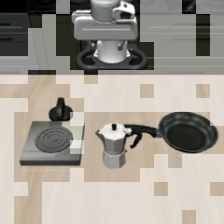}
&
\includegraphics[width=0.083\textwidth]{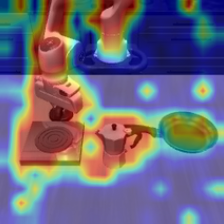}
\includegraphics[width=0.083\textwidth]{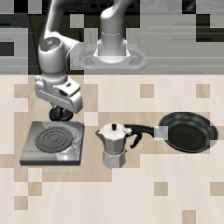}
&
\includegraphics[width=0.083\textwidth]{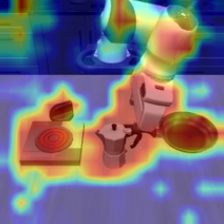}
\includegraphics[width=0.083\textwidth]{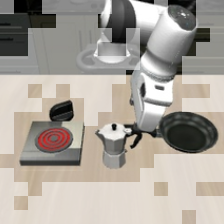}
&
\includegraphics[width=0.083\textwidth]{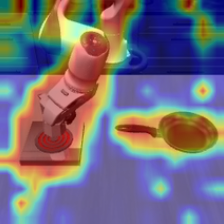}
\includegraphics[width=0.083\textwidth]{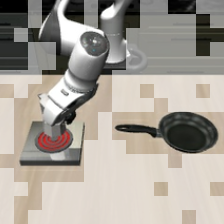}
\\[2pt]
\rotatebox{90}{\textbf{Wrist}}
&
\includegraphics[width=0.083\textwidth]{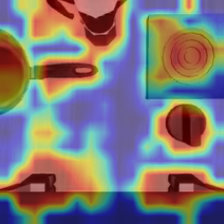}
\includegraphics[width=0.083\textwidth]{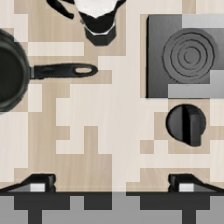}
&
\includegraphics[width=0.083\textwidth]{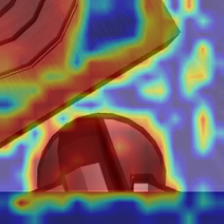}
\includegraphics[width=0.083\textwidth]{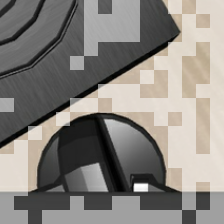}
&
\includegraphics[width=0.083\textwidth]{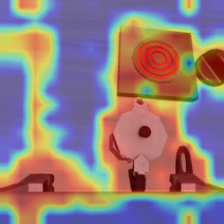}
\includegraphics[width=0.083\textwidth]{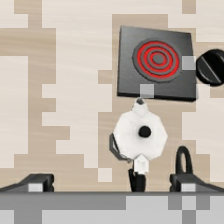}
&
\includegraphics[width=0.083\textwidth]{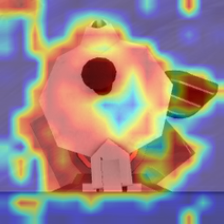}
\includegraphics[width=0.083\textwidth]{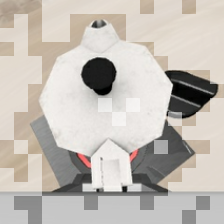}
\end{tabular}
\caption{
Dual-space attenuation dynamics on the LIBERO task \emph{``turn on the stove and put the moka pot on it.''}
For each timestep, the left image in each pair shows explicit relevance anchors $\alpha_i$, and the right image shows the corresponding latent attenuation map.
}
\label{fig:dual_space_vis}
\end{figure*}

\subsection{Analysis}

The ablations show that both relevance anchoring and dual-space attenuation are important. We further inspect whether the learned anchors and attenuation maps behave consistently with the intended mechanism and analyze the coupled budget--threshold configuration.

\paragraph{Dual-space attenuation dynamics.}
Figure~\ref{fig:dual_space_vis} compares the shared relevance signal in input and latent spaces. Across both views, high-relevance regions track the stove, moka pot, gripper, and interaction areas, while low-relevance regions remain visible but their corresponding visual hidden states are softly attenuated. Across LIBERO suites, mean anchors range from 0.36 to 0.46, with 41--54\% of visual tokens below the threshold.
Across time and views, emphasis shifts from object localization toward gripper--target interaction, indicating that the anchors track manipulation progress rather than static or viewpoint-specific saliency. Although their intensities differ after contextual computation, the spatial agreement between input anchors and latent maps shows that shared anchors maintain a task-conditioned reference as interaction context evolves.

\begin{center}
\begin{minipage}{\columnwidth}
\centering
\setlength{\tabcolsep}{6pt}
\begin{tabular*}{0.90\columnwidth}{@{\extracolsep{\fill}}ccc@{}}
\toprule
Budget $\rho$ & Threshold $\theta$ & Avg. \\
\midrule
0.2 & 0.1 & 96.0 \\
\textbf{0.4} & \textbf{0.3} & \textbf{98.0} \\
0.6 & 0.5 & 97.8 \\
0.8 & 0.7 & 97.0 \\
\bottomrule
\end{tabular*}
\captionof{table}{Sensitivity to coupled budget--threshold settings on LIBERO (average success rate, \%).}
\label{tab:budget_sensitivity}
\end{minipage}
\end{center}

\paragraph{Budget--threshold calibration.}
The relevance budget $\rho$ controls the overall relevance mass retained by the anchors, while the threshold $\theta$ determines which visual hidden states undergo latent-space attenuation. We therefore calibrate them jointly and set $\theta=\rho-0.1$ to maintain a consistent operating margin and stable training (Table~\ref{tab:budget_sensitivity}). The $0.4/0.3$ setting performs best at 98.0\%. Reducing the pair to $0.2/0.1$ lowers performance to 96.0\%, suggesting that excessive attenuation removes useful spatial context. Increasing it to $0.6/0.5$ causes only a minor drop to 97.8\%, whereas $0.8/0.7$ reaches 97.0\% as more distracting evidence is retained. This asymmetric degradation indicates that DIVA is more sensitive to suppressing useful context than to preserving modest residual distraction. Overall, the $0.4/0.3$ setting provides the best balance between retaining task-relevant visual evidence and attenuating irrelevant information across varied scenes.

\section{Conclusion}

We introduced DIVA, an anchor-then-attenuate framework that combines intent-conditioned relevance anchoring with dual-space attenuation while preserving the full visual token sequence. Grounded Text Pooling integrates task intent with localized visual evidence, and shared anchors regulate how strongly representations enter and persist through policy computation. Across LIBERO, zero-shot LIBERO-Plus, and physical tasks, DIVA improves OpenVLA-OFT under standard and perturbed conditions, while ablations confirm the complementary roles of anchoring and dual-space attenuation. These results support non-destructive visual-state strength control as a practical design for robust VLA policies.

\par\vspace{6pt}
\noindent\textbf{Limitations.}
Fixed budgets may be suboptimal as the amount and extent of task-relevant evidence vary across scenes. A single global value cannot match observations that differ substantially in object scale, clutter, or the spatial extent of useful evidence. With a fixed budget, small targets may receive unnecessarily broad support, whereas large objects or extended interaction regions may be underrepresented. This mismatch can introduce mild optimization noise and makes one global setting unlikely to be optimal for every observation. Future work will explore input-adaptive budgets that infer the appropriate relevance mass for each observation while retaining the stability of the current regularizer.

\clearpage
\bibliography{references}

\clearpage
\appendix
\setcounter{secnumdepth}{2}
\section*{Appendix}
\input{appendix}

\end{document}

%% file: appendix.tex
\section{Implementation Details}
\label{sec:implementation}

\noindent\textbf{Base OpenVLA-OFT architecture.~}
DIVA is instantiated on OpenVLA-OFT. The base policy uses a shared SigLIP--DINOv2 visual encoder for multi-view observations and projects the resulting visual features into the Llama-2 7B language-embedding space. A separate projector embeds robot proprioception, and an MLP action head predicts continuous action chunks. OpenVLA-OFT replaces the causal attention used by standard OpenVLA with bidirectional attention, allowing all actions in a chunk to be decoded in parallel. We retain this base architecture and its continuous L1 action objective.

\noindent\textbf{DIVA framework modifications.~}
DIVA adds an anchor-then-attenuate pathway to the projected visual-token sequence. Grounded Text Pooling first produces a compact task-intent representation by weighting instruction tokens according to localized visual support. We concatenate this task representation with each projected visual token and process the task-conditioned sequence using one self-attention block with a hidden width of 512 and eight attention heads. An independent sigmoid head then predicts a scalar relevance anchor for every visual patch, allowing all patches belonging to an object or interaction region to receive high relevance simultaneously.

The same anchors control visual representations in two complementary spaces. Before backbone entry, explicit input-space attenuation scales each projected visual token by its anchor and thereby initializes its representation strength. Within the language-action backbone, implicit latent-space attenuation reuses the anchors at every layer. Hidden states above the relevance threshold remain unchanged, whereas states below the threshold are softly interpolated toward their anchor-scaled values. This persistent intervention maintains attenuation after attention, feed-forward transformations, and residual updates while preserving the full visual-token sequence.

\noindent\textbf{Training details.~}
The main paper describes the matched training budgets and the timing of DIVA activation. Stage-specific settings are detailed below.

\noindent\textbf{LIBERO simulation.~}
We jointly train one policy on the four LIBERO suites. Stage 1 follows the standard OpenVLA-OFT recipe for 150K steps, using continuous L1 action regression, third-person and wrist observations, proprioception, image augmentation, and LoRA with rank 32. The action chunk length is 8, the effective batch size is 64, and training uses eight NVIDIA L20 GPUs. Stage 2 initializes from the 150K checkpoint, activates Grounded Text Pooling, relevance anchoring, and dual-space attenuation, and continues training for 10K steps. We use AdamW with a peak learning rate of $5\times10^{-4}$, a 1K-step linear warmup, cosine decay to $5\times10^{-5}$, and gradient clipping with global norm 1.0.

\noindent\textbf{Physical robot.~}
The physical-robot policies use continuous L1 action regression, third-person and wrist observations, proprioception, image augmentation, and LoRA rank 32, with an action chunk length of 25. Stage 1 trains OpenVLA-OFT for 20K steps. Stage 2 initializes from this checkpoint and trains DIVA for another 10K steps with the same warmup, cosine schedule, and gradient clipping used in simulation. Training uses four NVIDIA L20 GPUs with a per-GPU batch size of 4 and two gradient-accumulation steps, giving a global batch size of 32.

\noindent\textbf{DIVA hyperparameters.~}
Table~\ref{tab:diva_hparams} lists the method-specific settings corresponding to the notation in the main paper. Grounded Text Pooling evaluates localized visual support using temperatures $\tau_v$ and $\tau_t$ and excludes function words from the pooling set. The relevance estimator consists of the single self-attention block described above followed by an independent sigmoid prediction for each visual patch.

\begin{table}[t]
\centering
\small
\setlength{\tabcolsep}{5pt}
\begin{tabular}{@{}lll@{}}
\toprule
\textbf{Component} & \textbf{Setting} & \textbf{Value} \\
\midrule
\multirow{2}{*}{Grounded Text Pooling}
& Visual temperature $\tau_v$ & 0.10 \\
& Text temperature $\tau_t$ & 0.05 \\
\midrule
\multirow{3}{*}{Relevance estimator}
& Hidden width & 512 \\
& Attention heads & 8 \\
& Self-attention blocks & 1 \\
\midrule
\multirow{2}{*}{Budget objective}
& Target relevance $\rho$ & 0.40 \\
& Loss weight $\eta$ & 0.01 \\
\midrule
\multirow{2}{*}{Latent attenuation}
& Threshold $\theta$ & 0.30 \\
& Strength $\lambda$ & 0.50 \\
\bottomrule
\end{tabular}
\caption{DIVA hyperparameters used in the reported experiments.}
\label{tab:diva_hparams}
\end{table}

The input-space path multiplies each projected visual token by its anchor. In every backbone layer, the latent path leaves states with $\alpha_i\geq\theta$ unchanged and scales states below the threshold by $1-\lambda(1-\alpha_i)$. The same anchors are shared by the two paths. For multi-view observations, the relevance-budget loss is computed separately for each camera view and then averaged.

\noindent\textbf{Training initialization and continued training.~}
The two-stage recipe separates learning a stable visuomotor policy from learning how DIVA should regulate its visual representations. DIVA is initialized from a task-trained OpenVLA-OFT checkpoint, which stabilizes the relevance anchors and the coupled budget--threshold configuration. The baseline is trained for the same total budget, namely 160K steps in simulation and 30K steps on the physical robot, and all ablation variants are trained under the same matched budget.

\section{Real-World Experiment Details}
\label{sec:realworld}

We conduct the physical experiments on a UFACTORY UF850 arm. Demonstrations are collected through VR teleoperation from two synchronized Intel RealSense RGB cameras: a wrist-mounted view and a fixed third-person view. Both cameras record at $1920\times1080$ and 30~FPS, while robot transitions are logged at a nominal control rate of 10~Hz. The policy receives the two RGB views together with robot proprioception.

\subsection{Dataset Preparation}

The raw demonstrations preserve the two camera images and robot state at every transition. Before training, all four task datasets are converted to the OpenVLA-OFT-compatible RLDS representation. We center-crop each camera stream to a square field of view and resize it to $224\times224$. Each RLDS transition contains the two RGB observations, six joint angles, a 7D relative action, and the language instruction. The action consists of TCP translation, wrapped Euler-angle rotation, and a binary gripper command. A no-op filter removes transitions with negligible translation and rotation while always retaining gripper changes.

\subsection{Real-World Task Suites}

We evaluate four task suites spanning simple pick-and-place, language-conditioned object selection, and long-horizon action sequencing. In the instruction templates below, $X$, $Y$, and $Z$ denote object names supplied by the current task instance.

\noindent\textbf{Pick the instructed cube.~}
\begin{itemize}
    \item \textbf{Instruction:} ``Put $X$ in the cup,'' where $X$ is the instructed cube.
    \item \textbf{Task:} Identify the requested cube among three visually similar candidates, pick it, and place it into the cup.
    \item \textbf{Dataset:} 51 demonstrations.
\end{itemize}

\noindent\textbf{Stack the instructed cups.~}
\begin{itemize}
    \item \textbf{Instruction:} ``Stack $X$ on top of $Y$,'' where $X$ and $Y$ are the instructed source and target cups.
    \item \textbf{Task:} Bind both color expressions to the intended cups and stack the specified source cup on the target cup.
    \item \textbf{Dataset:} 51 demonstrations.
\end{itemize}

\noindent\textbf{Sequential button pressing.~}
\begin{itemize}
    \item \textbf{Instruction:} ``First press $X$, then press $Y$, and finally press $Z$,'' where $X$, $Y$, and $Z$ are the instructed colored buttons.
    \item \textbf{Task:} Press the three colored buttons in the instructed order while inferring progress from the current visual state.
    \item \textbf{Dataset:} 50 demonstrations.
\end{itemize}

\noindent\textbf{Color-conditioned placement.~}
\begin{itemize}
    \item \textbf{Instruction:} ``Place $X$ in $Y$,'' where $X$ is the manipulated object and $Y$ is the instructed colored bowl.
    \item \textbf{Task:} Pick the object and place it into the bowl specified by the instruction.
    \item \textbf{Dataset:} 51 demonstrations.
\end{itemize}

\subsection{Evaluation Conditions}

Every task is evaluated under three conditions. The \emph{Standard Setup} matches the training workspace. \emph{Unseen Objects} introduces task-irrelevant distractors near the interaction area, and \emph{Unseen Background} changes the tabletop appearance. We conduct 25 physical trials for every task--condition pair and report the number of successful executions. A trial succeeds only when the complete instructed task is finished within the evaluation horizon.

\subsection{Execution Trajectories}

\begin{figure*}[t]
\centering
\includegraphics[width=\textwidth]{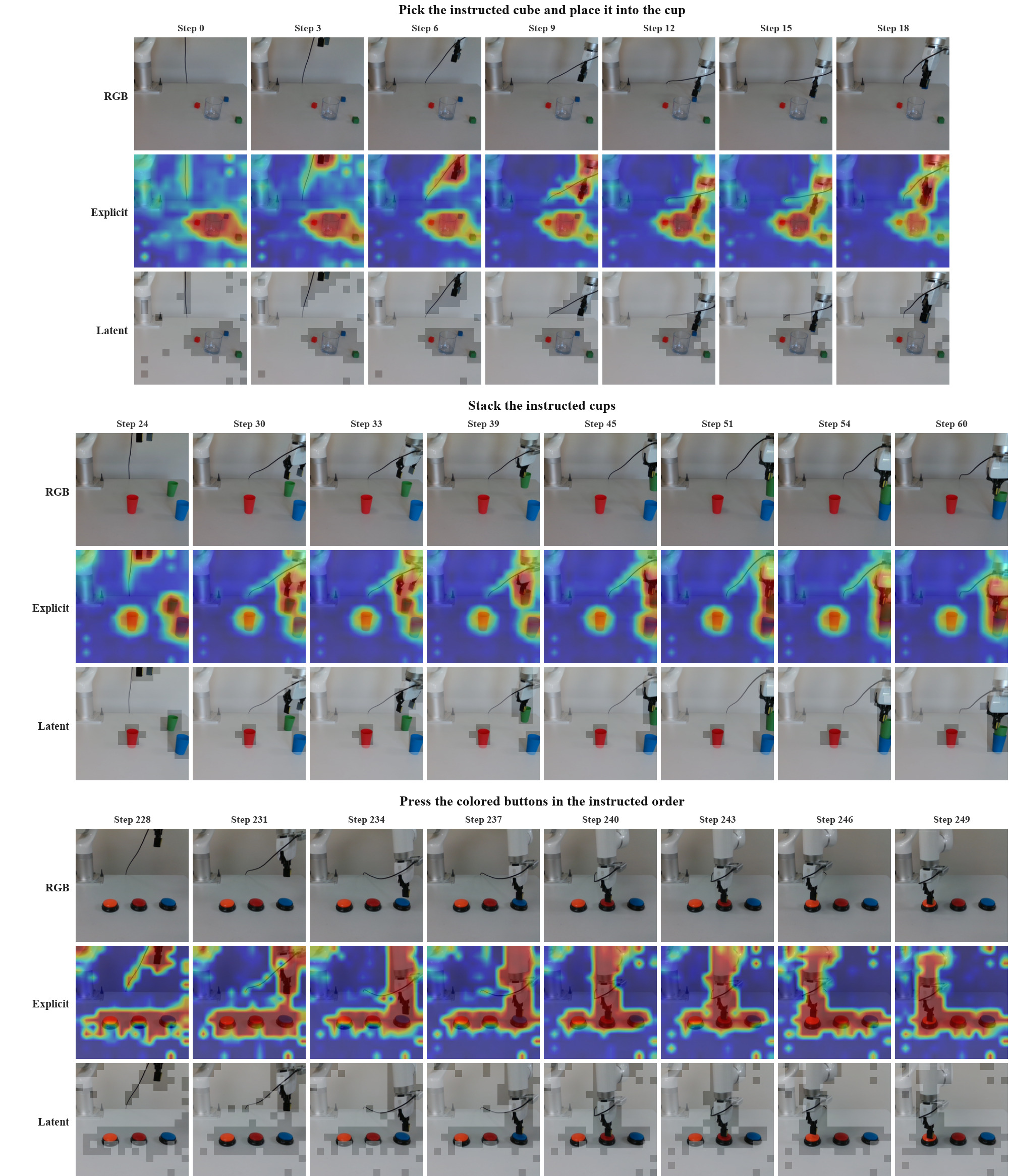}
\caption{Real-world trajectories under the standard setup for instructed-cube placement, instructed-cup stacking, and sequential button pressing. Each task is shown across all available saved queries, up to eight policy-query steps, with RGB, explicit, and latent attenuation rows.}
\label{fig:realworld_standard}
\end{figure*}

\begin{figure*}[t]
\centering
\includegraphics[width=\textwidth]{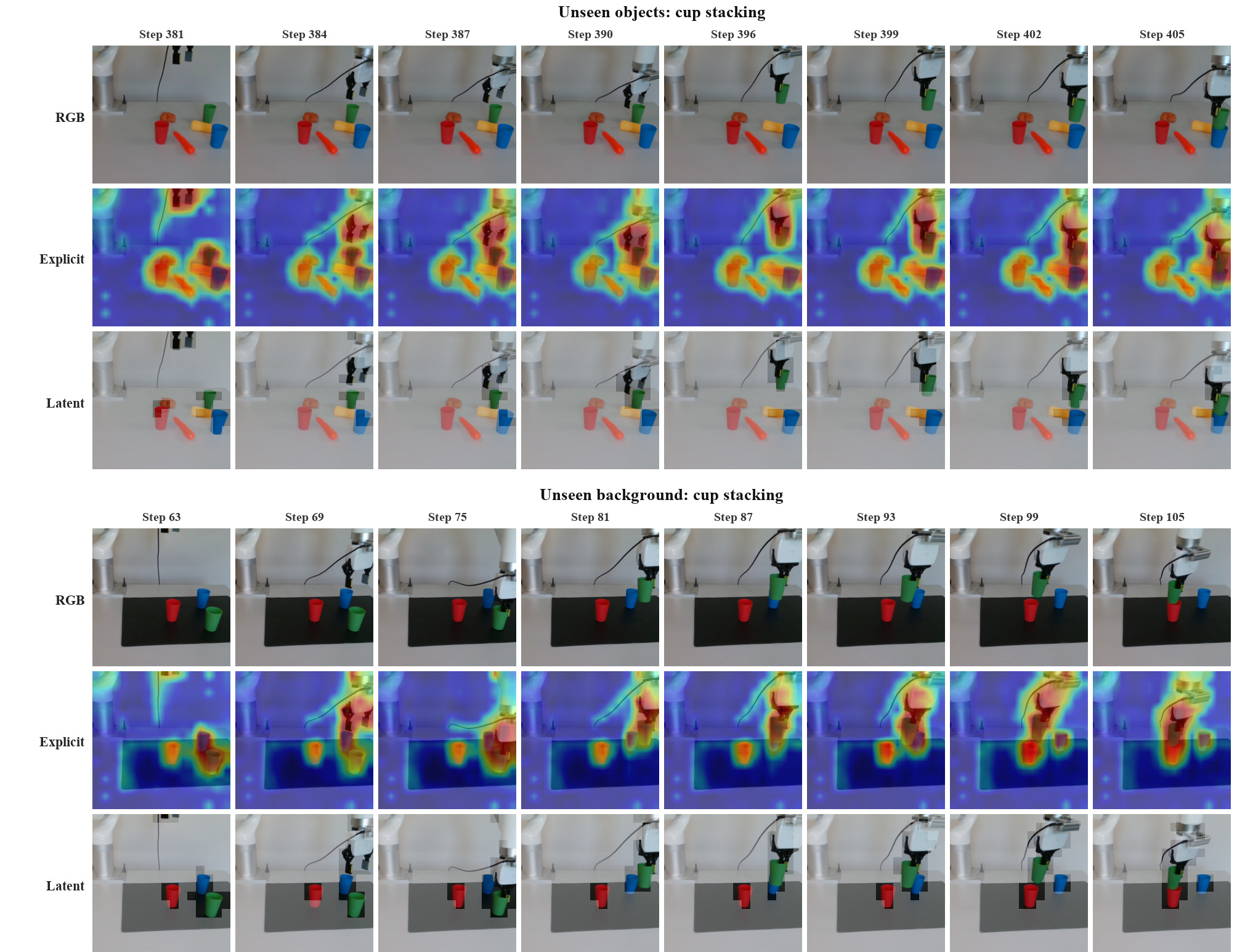}
\caption{Real-world cup-stacking trajectories under unseen object distractors and an unseen background. Each condition is shown across eight policy-query steps with RGB, explicit, and latent attenuation rows.}
\label{fig:realworld_ood_a}
\end{figure*}

Figures~\ref{fig:realworld_standard} and~\ref{fig:realworld_ood_a} provide representative execution trajectories from the primary third-person view. The standard trajectories cover the language-conditioned cube, cup-stacking, and button-sequencing tasks, while the out-of-distribution examples show cup stacking under unseen distractors and backgrounds.

\section{Simulation Benchmark Details}
\label{sec:evaluation}

\subsection{LIBERO}

LIBERO is a language-conditioned robot-manipulation benchmark built from procedurally generated tasks and human-teleoperated demonstrations. The complete benchmark contains 130 tasks: three controlled ten-task suites and LIBERO-100, a collection of 100 tasks with more entangled variation in objects, layouts, and behaviors. Following the standard VLA evaluation protocol, we use the 40-task split formed by LIBERO-Spatial, LIBERO-Object, LIBERO-Goal, and the ten long-horizon tasks from LIBERO-100:

\begin{itemize}
    \item \textbf{LIBERO-Spatial} keeps the objects and pick-and-place goal largely fixed while changing the spatial relation that identifies the target. For example, two visually identical bowls may be distinguished only by their position relative to other objects.
    \item \textbf{LIBERO-Object} retains a common pick-and-place structure while varying the manipulated object, isolating recognition and manipulation across object categories.
    \item \textbf{LIBERO-Goal} keeps the objects and their spatial arrangement fixed but changes the desired goal, requiring different motions and interaction behaviors in the same scene.
    \item \textbf{LIBERO-Long} contains ten long-horizon tasks selected from LIBERO-100. These tasks combine multiple object interactions and sequential manipulation stages.
\end{itemize}

The LIBERO training split, episode counts, image preprocessing, and success criterion are described in the main paper.

\subsection{LIBERO-Plus}

LIBERO-Plus extends the same four base suites with controlled visual, state, and language perturbations for robustness evaluation. It contains 10,030 task variants: 2,402 from LIBERO-Spatial, 2,518 from LIBERO-Object, 2,591 from LIBERO-Goal, and 2,519 from LIBERO-Long. Each variant is assigned to one of seven perturbation factors:

\begin{table}[t]
\centering
\small
\setlength{\tabcolsep}{8pt}
\begin{tabular}{@{}lr@{}}
\toprule
\textbf{Perturbation factor} & \textbf{Variants} \\
\midrule
Camera viewpoints & 1,599 \\
Robot initial states & 1,550 \\
Language instructions & 1,537 \\
Light conditions & 1,142 \\
Background textures & 1,076 \\
Sensor noise & 1,601 \\
Object layouts & 1,525 \\
\midrule
\textbf{Total} & \textbf{10,030} \\
\bottomrule
\end{tabular}
\caption{Composition of the LIBERO-Plus evaluation set.}
\label{tab:liberoplus_composition}
\end{table}

The camera factor changes viewpoint position, orientation, or field of view; robot-state variants perturb the manipulator's initial pose. Language variants rewrite the instruction while preserving the intended task. Lighting changes intensity, direction, color, or shadows, whereas background variants alter scene and surface textures. Sensor-noise variants apply photometric distortions or image degradation. Object-layout variants introduce confounding objects or displace target objects.

We follow the official zero-shot protocol: the policy is trained only on standard LIBERO demonstrations and receives no LIBERO-Plus perturbation data for training or adaptation. In each evaluation run, every variant is executed once. We report success separately for each factor and compute the overall score across all 10,030 variants; consequently, the total is task-weighted rather than an unweighted mean of the seven factor scores. The main paper additionally reports the mean and sample standard deviation over five repeated evaluations.

\section{Additional Qualitative Results}
\label{sec:qualitative}

The following visualizations show representative DIVA trajectories from the primary third-person view. Following the trajectory layout of prior VLA analyses, each task is displayed across up to eight evenly spaced policy-query steps. The three rows show the original RGB observations, explicit input-space attenuation, and corresponding latent attenuation results. Warm colors in the explicit maps indicate larger relevance anchors, while lightened patches indicate attenuated latent visual states.

\paragraph{Reading the trajectories.}
The sampled steps span the beginning, approach, interaction, and later stages of each rollout, allowing the evolution of the same task-conditioned relevance signal to be followed over time. The RGB row provides the unmodified observation. The Explicit row overlays the patch-wise anchors applied before backbone entry. Since latent hidden states are not directly image-valued, the Latent row spatially renders the attenuation applied to their corresponding visual-token positions: lighter patches receive stronger attenuation, whereas unmasked regions retain their original strength.

\paragraph{Qualitative progression.}
At the beginning of a trajectory, relevance is commonly distributed over candidate objects, the target region, the robot gripper, and nearby spatial context. As the policy approaches and interacts with the scene, the response becomes concentrated around the manipulated object and the active contact region. Long-horizon tasks exhibit a corresponding shift between successive subgoals. In the real-world OOD examples, unseen objects and background changes alter the visual appearance substantially, yet the instructed objects and interaction regions remain emphasized while distractors are attenuated. Together, the sequences illustrate DIVA's task-conditioned and temporally evolving control signal.

\paragraph{Interpreting the two attenuation spaces.}
The two rows visualize different uses of the same anchors. Explicit attenuation sets the initial strength of projected visual tokens, whereas latent attenuation repeatedly acts only on states below the threshold. The latent view therefore appears more selective because it shows positions receiving persistent attenuation rather than a second relevance prediction. Across trajectories, the manipulated object, target, gripper, and active contact region remain consistently retained.

\paragraph{Behavior under visual shifts.}
With unseen distractors and backgrounds, DIVA continues to preserve the instructed cups, gripper, and interaction region while weakening surrounding visual changes. The retained regions also evolve with execution: early steps cover candidate objects and spatial context, while later steps emphasize the manipulated object and its destination. This progression shows that attenuation follows both task identity and execution stage.

\begin{figure*}[t]
\centering
\includegraphics[width=\textwidth]{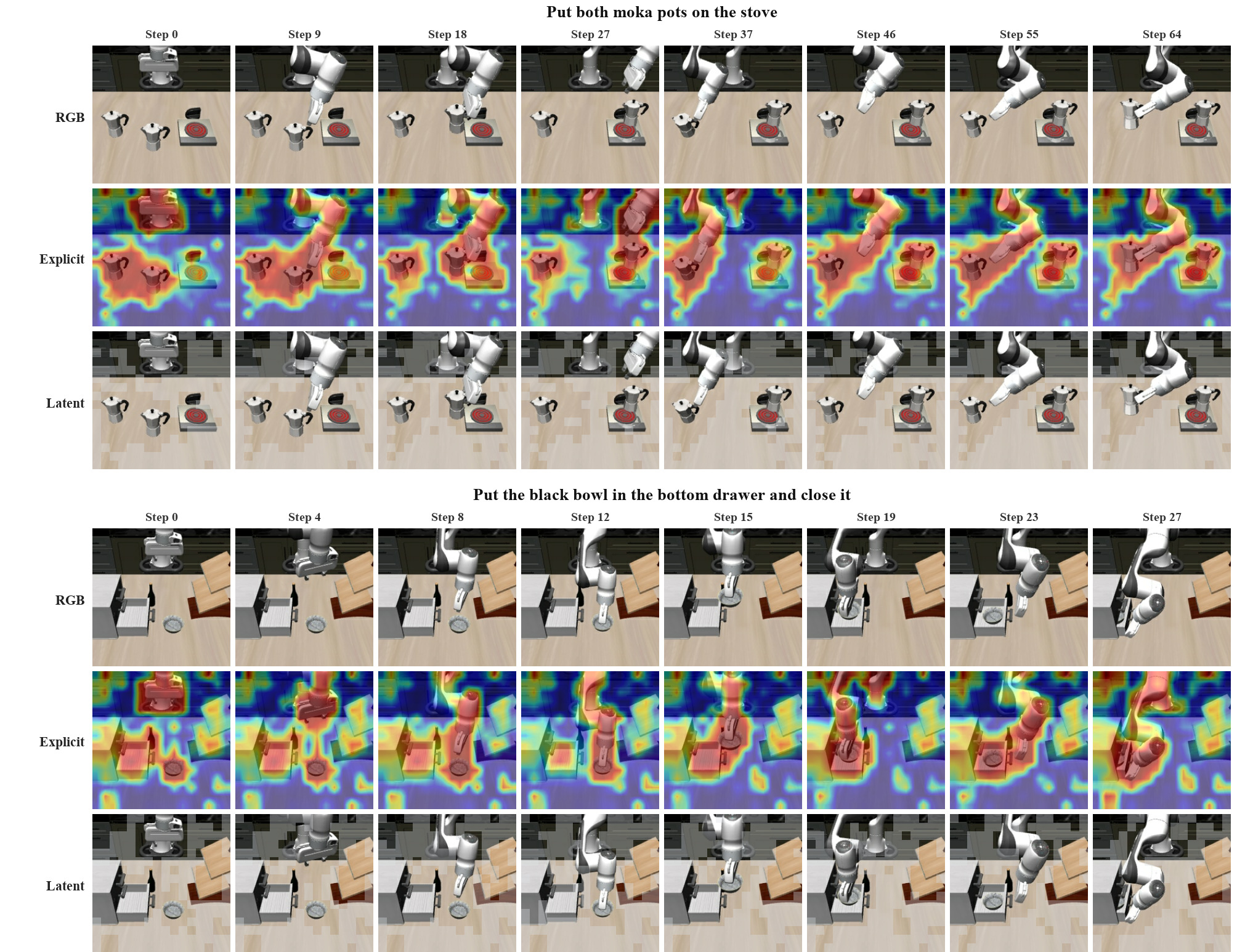}
\caption{DIVA dynamics on two LIBERO long-horizon tasks across eight policy-query steps. Each task uses rows for RGB, relevance, and latent attenuation.}
\label{fig:simulation_trajectories_a}
\end{figure*}

\begin{figure*}[t]
\centering
\includegraphics[width=\textwidth]{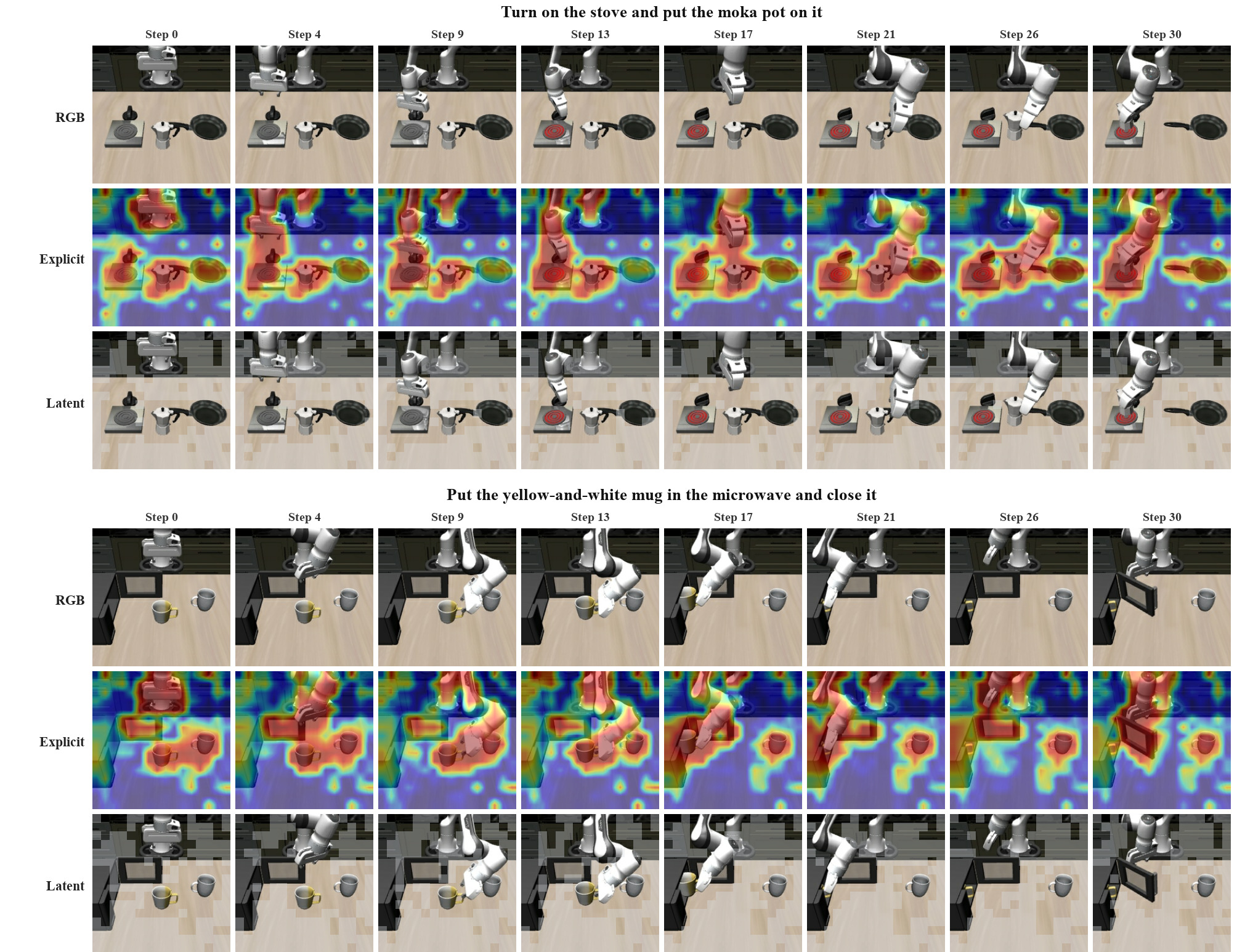}
\caption{DIVA dynamics on two additional LIBERO tasks across eight policy-query steps. Relevance shifts from candidate objects and targets toward the manipulated object and active interaction region as execution progresses.}
\label{fig:simulation_trajectories_b}
\end{figure*}

